# Conditional Plausibility Measures and Bayesian Networks

**Joseph Y. Halpern**
Dept. Computer Science
Cornell University
Ithaca, NY 14853
halpern@cs.cornell.edu
http://www.cs.cornell.edu/home/halpern

## Abstract

A general notion of algebraic conditional plausibility measures is defined. Probability measures, ranking functions, possibility measures, and (under the appropriate definitions) sets of probability measures can all be viewed as defining algebraic conditional plausibility measures. It is shown that the technology of Bayesian networks can be applied to algebraic conditional plausibility measures.

## 1 INTRODUCTION

Pearl [1988] among others has long argued that Bayesian networks (that is, the dags without the conditional probability tables) represent important *qualitative* information about uncertainty regarding conditional dependencies and independencies. To the extent that this is true, Bayesian networks should make perfect sense for non-probabilistic representations of uncertainty. And, indeed, Bayesian networks have been used with $\kappa$ rankings [Spohn 1988] by Darwiche and Goldszmidt [1994]. It follows from results of Wilson [1994] that the technology of Bayesian networks can also be applied to *possibility measures* [Dubois and Prade 1990].

The question I address in this paper is "What properties of a representation of uncertainty are required in order for the technology of Bayesian networks to work?" This question too has been addressed in earlier work, see [Darwiche 1992; Darwiche and Ginsberg 1992; Friedman and Halpern 1995; Wilson 1994], although the characterization given here is somewhat different. Here I represent uncertainty using *plausibility measures*, as in [Friedman and Halpern 1995]. To answer the question, I must examine general properties of conditional plausibility as well as defining a notion of plausibilistic independence. Unlike earlier papers, I enforce a symmetry condition in the definition of conditional independence, so that, for example, $A$ is independent of $B$ iff $B$ is independent of $A$. While this property holds for probability, under the asymmetric definition of independence used in earlier work it does not necessarily hold for other formalisms. There are also subtle but important differences between this paper and [Friedman and Halpern 1995] in the notion of conditional plausibility. The definitions here are simpler but more general; particular attention is paid here to conditions on when the conditional plausibility must be defined.

The major results here are a general condition, simpler than that given in [Friedman and Halpern 1995; Wilson 1994], under which a conditional plausibility measure satisfies the semi-graphoid properties (which means it can be represented using a Bayesian network). There is also a weak condition that suffices to guarantee that d-separation in the network characterizes conditional independence. The conditions clearly apply to $\kappa$ rankings and possibility measures. Perhaps more interestingly, they also apply to sets of probabilities under a novel representation of such sets as a plausibility measure. This novel representation (and the associated notion of conditioning) is shown to have some natural properties not shared by other representations.

The rest of the paper is organized as follows. In Section 2, I discuss conditional plausibility measures. Section 3 introduces *algebraic* conditional plausibility measures, which are ones where there is essentially an analogue to $+$ and $\times$. (Putting such an algebraic structure on uncertainty is not new; it was also done in [Darwiche 1992; Darwiche and Ginsberg 1992; Friedman and Halpern 1995; Weydert 1994].) Section 4 discusses independence and conditional independence in conditional plausibility spaces, and shows that algebraic conditional plausibility measures satisfy the semi-graphoid properties. Finally, in Section 5, Bayesian networks based on (algebraic) plausibility measures are considered. Combining the fact that algebraic plausibility measures satisfy the semi-graphoid properties with the results of [Geiger, Verma, and Pearl 1990], it follows that d-separation in a Bayesian network $G$ implies conditional independence for all algebraic plausibility measures compatible with $G$; a weak richness condition is shown to yield the converse. The paper concludes in Section 6. For reasons of space, proofs are omitted; they can be found in the full paper.

## 2 CONDITIONAL PLAUSIBILITY

The basic idea behind plausibility measures is straightforward. A probability measure maps subsets of a set $W$ to $[0, 1]$. Its domain may not consist of all subsets of $W$; however, it is required to be an *algebra*. (Recall that an algebra $\mathcal{F}$ over $W$ is a set of subsets of $W$ containing $W$ and closed under union and complementation, so that if $U, V \in \mathcal{F}$, then so are $U \cup V$ and $\overline{U}$.) A *plausibility measure* is more general; it maps elements in an algebra $\mathcal{F}$ to some arbitrary partially ordered set. If Pl is a plausibility measure, then we read $\text{Pl}(U)$ as "the plausibility of set $U$". If $\text{Pl}(U) \leq \text{Pl}(V)$, then $V$ is at least as plausible as $U$. Because the ordering is partial, it could be that the plausibility of two different sets is incomparable. An agent may not be prepared to say of two sets that one is more likely than another or that they are equal in likelihood.

Formally, a *plausibility space* is a tuple $S = (W, \mathcal{F}, \text{Pl})$, where $W$ is a set of worlds, $\mathcal{F}$ is an algebra over $W$, and Pl maps sets in $\mathcal{F}$ to some set $D$ of *plausibility values* partially ordered by a relation $\leq_D$ (so that $\leq_D$ is reflexive, transitive, and anti-symmetric) that contains two special elements $\top_D$ and $\bot_D$ such that $\bot_D \leq_D d \leq_D \top_D$ for all $d \in D$. In the case of probability measures, $\top_D$ and $\bot_D$ are 1 and 0, respectively. As usual, the ordering is defined $<_D$ by taking $d_1 <_D d_2$ if $d_1 \leq_D d_2$ and $d_1 \neq d_2$. I omit the subscript $D$ from $\leq_D, <_D, \top_D$ and $\bot_D$ whenever it is clear from context.

There are three requirements on plausibility measures. The first two are obvious analogues of requirements that hold for other notions of uncertainty: the whole space gets the maximum plausibility and the empty set gets the minimum plausibility. The third requirement says that a set must be at least as plausible as any of its subsets.

Pl1. $\text{Pl}(\emptyset) = \bot_D$.

Pl2. $\text{Pl}(W) = \top_D$.

Pl3. If $U \subseteq U'$, then $\text{Pl}(U) \leq \text{Pl}(U')$.

(In Pl3, I am implicitly assuming that $U, U' \in \mathcal{F}$. Similar assumptions are made throughout.)

All the standard representations of uncertainty in the literature can be represented as plausibility measures. I briefly describe some other representations of uncertainty that will be of relevance to this paper.

**Sets of probabilities:** One common way of representing uncertainty is by a set of probability measures. This set is often assumed to be convex (see, for example, [Campos and Moral 1995; Couso, Moral, and Walley 1999; Walley 1991] and the references therein), however, convex sets do not seem appropriate for representing independence assumptions, so I do not make this restriction here. For example, if a coin with an unknown probability of heads is tossed twice, and the tosses are known to be independent, it seems that a reasonable representation is given by the set $\mathcal{P}_0$ consisting of all measures $\mu_\alpha$, where $\mu_\alpha(hh) = \alpha^2$, $\mu_\alpha(ht) = \mu_\alpha(th) = \alpha(1-\alpha), \mu_\alpha(tt) = (1-\alpha)^2$. Unfortunately, $\mathcal{P}_0$ is not convex. Moreover, its convex hull includes many measures for which the coin tosses are not independent. It is argued in [Couso, Moral, and Walley 1999] that a set of probability measures is behaviorally equivalent to its convex hull. However, even if we accept this argument, it does not follow that a set and its convex hull are equivalent insofar as determination of independencies goes.

There are a number of ways of viewing a set $\mathcal{P}$ of probability measures as a plausibility measure. One uses the *lower probability* $\mathcal{P}_*$, defined as $\mathcal{P}_*(U) = \inf\{\mu(U) : \mu \in \mathcal{P}\}$. Clearly $\mathcal{P}_*$ satisfies Pl1–3. The corresponding *upper probability* $P^*$, defined as $\mathcal{P}^*(U) = \sup\{\mu : \mu \in \mathcal{P}\} = 1 - \mathcal{P}_*(U)$, is also clearly a plausibility measure.

Both $\mathcal{P}_*$ and $\mathcal{P}^*$ give a way of comparing the likelihood of two subsets $U$ and $V$ of $W$. These two ways are incomparable; it is easy to find a set $\mathcal{P}$ of probability measures on $W$ and subsets $U$ and $V$ of $W$ such that $\mathcal{P}_*(U) < \mathcal{P}_*(V)$ and $\mathcal{P}^*(U) > \mathcal{P}^*(V)$. Rather than choosing between $\mathcal{P}_*$ and $\mathcal{P}^*$, we can associate a different plausibility measure with $\mathcal{P}$ that captures both. Let $D_{\mathcal{P}_*, \mathcal{P}^*} = \{(a, b) : 0 \leq a \leq b \leq 1\}$ and define $(a, b) \leq (a', b')$ iff $b \leq a'$. This puts a partial order on $D_{\mathcal{P}_*, \mathcal{P}^*}$, with $\bot_{D_{\mathcal{P}_*, \mathcal{P}^*}} = (0, 0)$ and $\top_{D_{\mathcal{P}_*, \mathcal{P}^*}} = (1, 1)$. Define $\text{Pl}_{\mathcal{P}_*, \mathcal{P}^*}(U) = (\mathcal{P}_*(U), \mathcal{P}^*(U))$. Thus, $\text{Pl}_{\mathcal{P}_*, \mathcal{P}^*}$ associates with a set $U$ two numbers which can be thought of as defining an interval in terms of the lower and upper probability of $U$. It is easy to check that $\text{Pl}_{\mathcal{P}_*, \mathcal{P}^*}$ satisfies Pl1–3, so it is indeed a plausibility measure, but one which puts only a partial order on events.

The trouble with $\mathcal{P}_*$, $\mathcal{P}^*$, and even $\text{Pl}_{\mathcal{P}_*, \mathcal{P}^*}$ is that they lose information. For example, it is not hard to find a set $\mathcal{P}$ of probability measures and subsets $U, V$ of $W$ such that $\mu(U) \leq \mu(V)$ for all $\mu \in \mathcal{P}$ and $\mu(U) < \mu(V)$ for some $\mu \in \mathcal{P}$, but $\mathcal{P}_*(U) = \mathcal{P}_*(V)$ and $\mathcal{P}^*(U) = \mathcal{P}^*(V)$. Indeed, there exists an infinite set $\mathcal{P}$ of probability measures such that $\mu(U) < \mu(V)$ for all $\mu \in \mathcal{P}$ but $\mathcal{P}_*(U) = \mathcal{P}_*(V)$ and $\mathcal{P}^*(U) = \mathcal{P}^*(V)$. If all the probability measures in $\mathcal{P}$ agree that $U$ is less likely than $V$, it seems reasonable to conclude that $U$ is less likely than $V$. However, none of $\mathcal{P}_*, \mathcal{P}^*$, or $\text{Pl}_{\mathcal{P}_*, \mathcal{P}^*}$ will necessarily draw this conclusion.

Fortunately, it is not hard to associate yet another plausibility measure with $\mathcal{P}$ that does not lose this important information. Let $D_\mathcal{P} = [0, 1]^\mathcal{P}$ with the pointwise ordering, so that $f \leq g$ iff $f(\mu) \leq g(\mu)$ for all $\mu \in \mathcal{P}$. Note that $\bot_{D_\mathcal{P}}$ is the function $f : \mathcal{P} \to [0, 1]$ such that $f(\mu) = 0$ for all $\mu \in \mathcal{P}$ and $\top_{D_\mathcal{P}}$ is the function $g$ such that $g(\mu) = 1$ for all $\mu \in \mathcal{P}$. For $U \subseteq W$, let $f_U$ be the function such that $f_U(\mu) = \mu(U)$ for all $\mu \in \mathcal{P}$. For example, for the set $\mathcal{P}_0$ of measures representing the two coin tosses, the set $W$ taken to be $\{hh, ht, tt, th\}$. Then, for example $f_{\{hh\}}(\mu_\alpha) = \mu_\alpha(hh) = \alpha^2$ and $f_{\{ht, tt\}}(\mu_\alpha) = 1 - \alpha$.

It is easy to see that, in general, $f_\emptyset = \bot_{D_\mathcal{P}}$ and $f_W = \top_{D_\mathcal{P}}$. Now define $\text{Pl}_\mathcal{P}(U) = f_U$. Thus, $\text{Pl}_\mathcal{P}(U) \leq \text{Pl}_\mathcal{P}(V)$ iff $f_U(\mu) \leq f_V(\mu)$ for all $\mu \in \mathcal{P}$ iff $\mu(U) \leq \mu(V)$ for all $\mu \in \mathcal{P}$. Clearly $\text{Pl}_\mathcal{P}$ satisfies Pl1–3. Pl1 and Pl2 follow since $\text{Pl}_\mathcal{P}(\emptyset) = f_\emptyset = \bot_{D_\mathcal{P}}$ and $\text{Pl}_\mathcal{P}(W) = f_W = \top_{D_\mathcal{P}}$, while Pl3 follows since if $U \subseteq V$ then $\mu(U) \leq \mu(V)$ for all $\mu \in \mathcal{P}$. $\text{Pl}_\mathcal{P}$ captures all the information in $\mathcal{P}$ (unlike, say, $\mathcal{P}_*$, which washes much of it away by taking infs).

This way of associating a plausibility measure with a set $\mathcal{P}$ of probability measures generalizes: it provides an approach to associating a single plausibility measure with any set of plausibility measures; I leave the straightforward details to the reader.

**Possibility measures:** A *possibility measure* Poss on $W$ is a function mapping subsets of $W$ to $[0, 1]$ such that $\text{Poss}(W) = 1$, $\text{Poss}(\emptyset) = 0$, and $\text{Poss}(U) = \sup_{w \in U}(\text{Poss}(\{w\}))$, so that $\text{Poss}(U \cup V) = \max(\text{Poss}(U), \text{Poss}(V))$ [Dubois and Prade 1990]. Clearly a possibility measure is a plausibility measure.

**Ranking functions:** An *ordinal ranking* (or $\kappa$-*ranking* or *ranking function*) $\kappa$ on $W$ (as defined by [Goldszmidt and Pearl 1992], based on ideas that go back to [Spohn 1988]) is a function mapping subsets of $W$ to $\mathbb{N}^* = \mathbb{N} \cup \{\infty\}$ such that $\kappa(W) = 0$, $\kappa(\emptyset) = \infty$, and $\kappa(U) = \min_{w \in U}(\kappa(\{w\}))$, so that $\kappa(U \cup V) = \min(\kappa(U), \kappa(V))$. Intuitively, a ranking function assigns a degree of surprise to each subset of worlds in $W$, where 0 means unsurprising and higher numbers denote greater surprise. It is easy to see that if $\kappa$ is a ranking function on $W$, then $(W, 2^W, \kappa)$ is a plausibility space, where $x \leq_{\mathbb{N}^*} y$ if and only if $y \leq x$ under the usual ordering on the ordinals. One standard view of a ranking function, going back to Spohn, is that a ranking of $k$ can be associated with a probability of $\epsilon^k$, for some fixed (possibly infinitesimal) $\epsilon$. Note that this viewpoint justifies taking $\kappa(W) = 0$, $\kappa(\emptyset) = \infty$, and $\kappa(U \cup V) = \min(\kappa(U), \kappa(V))$.

Since Bayesian networks make such heavy use of conditioning, my interest here is not just plausibility measures, but *conditional* plausibility measures (*cpm*'s). Given a set $W$ of worlds, a cpm maps pairs of subsets of $W$ to some partially ordered set $D$. I write $\text{Pl}(U|V)$ rather than $\text{Pl}(U, V)$, in keeping with standard notation for conditioning. In the case of a probability measure $\mu$, it is standard to take $\mu(U|V)$ to be undefined in $\mu(V) = 0$. In general, we must make precise what the allowable second arguments are. Thus, I take the domain of a cpm to have the form $\mathcal{F} \times \mathcal{F}'$ where, intuitively, $\mathcal{F}'$ consists of those sets in $\mathcal{F}$ on which it makes sense to condition. For example, if we start with an unconditional probability measure $\mu$, $\mathcal{F}'$ might consist of all sets $V$ such that $\mu(V) > 0$. (Note that $\mathcal{F}'$ is not an algebra—it is not closed under either intersection or complementation.) A *Popper algebra* over $W$ is a set $\mathcal{F} \times \mathcal{F}'$ of subsets of $W \times W$ satisfying the following properties:

Acc1. $\mathcal{F}$ is an algebra over $W$.

Acc2. $\mathcal{F}'$ is a nonempty subset of $\mathcal{F}$.

Acc3. $\mathcal{F}'$ is closed under supersets in $\mathcal{F}$, in that if $V \in \mathcal{F}'$, $V \subseteq V'$, and $V' \in \mathcal{F}$, then $V' \in \mathcal{F}'$.

(Popper algebras are named after Karl Popper, who was the first to consider conditional probability as the basic notion [Popper 1968].)

A *conditional plausibility space* (*cps*) is a tuple $(W, \mathcal{F}, \mathcal{F}', \text{Pl})$, where $\mathcal{F} \times \mathcal{F}'$ is a Popper algebra over $W$, $\text{Pl} : \mathcal{F} \times \mathcal{F}' \to D$, $D$ is a partially ordered domain of plausibility values, and Pl is a *conditional plausibility measure* (cpm) that satisfies the following conditions:

CPl1. $\text{Pl}(\emptyset|V) = \perp_D$.

CPl2. $\text{Pl}(W|V) = \top_D$.

CPl3. If $U \subseteq U'$, then $\text{Pl}(U|V) \leq \text{Pl}(U'|V)$.

CPl4 $\text{Pl}(U|V) = \text{Pl}(U \cap V|V)$.

CPl1–3 are the obvious analogues to Pl1–3. CPl4 is a minimal property that guarantees that when conditioning on $V$, everything is relativized to $V$. It follows easily from CPl1–4 thatx $\text{Pl}(\cdot|V)$ is a plausibility measure on $V$ for each fixed $V$. A cps is *acceptable* if it satisfies

Acc4. If $V \in \mathcal{F}'$, $U \in \mathcal{F}$, and $\text{Pl}(U|V) \neq \perp_D$, then $U \cap V \in \mathcal{F}'$.

Acceptability is a generalization of the observation that if $\text{Pr}(V) \neq 0$, then then conditioning on $V$ should be defined. It says that if $\text{Pl}(U|V) \neq \perp_D$, then conditioning on $V \cap U$ should be defined.

This notion of cps is closely related to that defined in [Friedman and Halpern 1995]. There, a conditional plausibility space is defined to be a family $\{W, D_V, \text{Pl}_V) : V \subseteq W\}$ of plausibility spaces that satisfies the following coherence condition, which relates conditioning on two different sets, where $\mathcal{F} = 2^W$ and $\mathcal{F}' = 2^W - \{\emptyset\}$:

CPl5. If $V \cap V' \in \mathcal{F}'$ and $U, U' \in \mathcal{F}$, then $\text{Pl}(U|V \cap V') \leq \text{Pl}(U'|V \cap V')$ iff $\text{Pl}(U \cap V|V') \leq \text{Pl}(U' \cap V|V')$.

It is not hard to show that CPl5 implies CPl4. However, CPl5 does not follow from CPl1–4 (indeed, as shown below, the standard notion of conditioning for lower probabilities satisfies CPl1–4 but not CPl5). A cps that satisfies CPl5 is said to be *coherent*. Although I do not assume CPl5 here, it in fact holds for all plausibility measures to which one of the main results applies (see Lemma 3.2).

To distinguish the definition of cps given in this paper from that given in [Friedman and Halpern 1995], I call the latter an FH-cps. There is no analogue to Acc1–4 in [Friedman and Halpern 1995]; $\mathcal{F}$ is implicitly taken to be $2^W$, while $\mathcal{F}'$ is implicitly taken to be $2^W - \{\emptyset\}$. This is an inessential difference between the definitions. More significantly, note that in an FH-cps, $(W, D_V, \text{Pl}_V)$ is a plausibility space for each fixed $V$, and thus satisfies Pl1–3. However, requiring CPl1–3 is *a priori* stronger than requiring Pl1–3 for each separate plausibility space. Pl1 requires that $\text{Pl}(\emptyset|V) = \perp_{D_V}$, but the elements $\perp_{D_V}$ may be different for each $V$. By way of contrast, CPl1 requires that $\text{Pl}(\perp|V)$ must be the same element, $\perp_D$, for all $V$. Similar remarks hold for Pl2. Nevertheless, as I show in the full paper, there is a construction that converts an FH-cps to a coherent cps.

I now consider some standard ways of getting cps's, starting with the unconditional representations of uncertainty discussed earlier. A cpm Pl *extends* an unconditional plausibility measure $\text{Pl}'$ if $\text{Pl}(U|W) = \text{Pl}'(U)$. All the constructions given below result in extensions.

**Ranking functions:** Given an unconditional ranking function $\kappa$, there is a well-known way of extending it to a conditional ranking function:

$$\kappa(U|V) = \begin{cases} \kappa(U \cap V) - \kappa(V) & \text{if } \kappa(V) \neq \infty, \\ \text{undefined} & \text{if } \kappa(V) = \infty. \end{cases}$$

This is consistent with the view that if $\kappa(V) = k$, then $\mu(V) = \epsilon^k$, since then $\kappa(U|V) = \epsilon^{\kappa(U \cap V) - \kappa(V)}$. It is easy to check that this definition results in a coherent cps.

**Possibility measures:** There are two standard ways of defining a conditional possibility measure from an unconditional possibility measure Poss. To distinguish them, I write $\text{Poss}(U|V)$ for the first approach and $\text{Poss}(U||V)$ for the second approach. According to the first approach,

$$\text{Poss}(U|V) = \begin{cases} \text{Poss}(V \cap U) & \text{if } \text{Poss}(V \cap U) < \text{Poss}(V), \\ 1 & \text{if } \text{Poss}(V \cap U) = \text{Poss}(V) > 0, \\ \text{undefined} & \text{if } \text{Poss}(V) = 0. \end{cases}$$

The second approach looks more like conditioning in probability:

$$\text{Poss}(U||V) = \begin{cases} \text{Poss}(V \cap U)/\text{Poss}(V) & \text{if } \text{Poss}(V) > 0, \\ \text{undefined} & \text{if } \text{Poss}(V) = 0. \end{cases}$$

It is easy to show that both definitions result in coherent cps's. (Many other notions of conditioning for possibility measures can be defined; see, for example [Fonck 1994]. I focus on these two because they are the ones most-often considered in the literature.)

**Sets of probabilities:** For a set $\mathcal{P}$ of probabilities, conditioning can be defined for all the representations of $\mathcal{P}$ as a plausibility measure. But in each case there are subtle choices involving when conditioning is undefined. For example, one definition of conditional lower probability is that $\mathcal{P}_*(U|V)$ is $\inf\{\mu(U|V) : \mu(V) \neq 0\}$ if $\mu(V) \neq 0$ for all $\mu \in \mathcal{P}$, and is undefined otherwise (i.e., if $\mu(V) = 0$ for some $\mu \in \mathcal{P}$). It is easy to check that $\mathcal{P}_*$ defined this way gives a coherent cpm, as does the corresponding definition of $\mathcal{P}^*$. The problem with this definition is that it may result in a rather small set $\mathcal{F}'$ for which conditioning is defined. For example, if for each set $V \neq W$, there is some measure $\mu \in \mathcal{P}$ such that $\mu(V) = 0$ (which can certainly happen in some nontrivial examples), then $\mathcal{F}' = \{W\}$.

The following definition gives a lower probability which is defined on more arguments:

$$\mathcal{P}_*(U|V) = \begin{cases} \inf\{\mu(U|V) : \mu(V) \neq 0\} \\ \quad \text{if } \mu(V) \neq 0 \text{ for some } \mu \in \mathcal{P}, \\ \text{undefined} \\ \quad \text{if } \mu(V) = 0 \text{ for all } \mu \in \mathcal{P}. \end{cases}$$

It is easy to see that this definition agrees with the first one whenever the first is defined and results, in general, in a larger set $\mathcal{F}'$. However, the second definition does not satisfy CPl5. For example, suppose that $W = \{a, b, c\}$ and $\mathcal{P} = \{\mu, \mu'\}$, where $\mu(a) = \mu(b) = 0$, $\mu(c) = 1$, $\mu'(a) = 2/3$, $\mu'(b) = 1/3$, and $\mu'(c) = 0$. Taking $V = \{a, b\}$, $U = \{a\}$, and $U' = \{b\}$, it is easy to see that according to the second definition, $\mathcal{P}_*(U \cap V|W) = \mathcal{P}_*(U' \cap V|W) = 0$, but $\mathcal{P}_*(U|V) > \mathcal{P}_*(U'|V)$.

For $\text{Pl}_\mathcal{P}$, there are two analogous definitions. For the first, $\text{Pl}_\mathcal{P}(U|V)$ is defined only if $\mu(V) > 0$ for all $\mu \in \mathcal{P}$, in which case $\text{Pl}_\mathcal{P}(U|V)$ is $f_{U|V}$, where $f_{U|V}(\mu) = \mu(U|V)$. This definition gives a coherent cps, but again, the problem is that $\mathcal{F}'$ may be small. Thus, in this paper, I use the following more general approach.

First extend $D_\mathcal{P}$ by allowing functions which have value $*$ (intuitively, $*$ denotes undefined). More precisely, let $D'_\mathcal{P}$ consist of all functions $f$ from $\mathcal{P}$ to $[0,1] \cup \{*\}$ such that $f(\mu) \neq *$ for at least one $\mu \in \mathcal{P}$. The idea is to define $\text{Pl}_\mathcal{P}(U|V) = f_{U|V}$, where $f_{U|V}(\mu) = \mu(U|V)$ if $\mu(V) > 0$ and $*$ otherwise. (Note that this agrees with the previous definition, which applies only to the situation where $\mu(V) > 0$ for all $\mu \in \mathcal{P}$.) There is a problem though, one to which I have already alluded, CPl1 says that $f_{\emptyset|V}$ must be $\bot$ for all $V$. Thus, it must be the case that $f_{\emptyset|V_1} = f_{\emptyset|V_2}$ for all $V_1, V_2 \subseteq W$. But if $\mu \in \mathcal{P}$ and $V_1, V_2 \subseteq W$ are such that $\mu(V_1) > 0$ and $\mu(V_2) = 0$, then $f_{\emptyset|V_1}(\mu) = 0$ and $f_{\emptyset|V_2}(\mu) = *$, so $f_{\emptyset|V_1} \neq f_{\emptyset|V_2}$. A similar problem arises with CPl2.

To deal with this problem $D'_\mathcal{P}$ must be slightly modified. Say that $f \in D'_\mathcal{P}$ is *equivalent to* $\bot_{D^*_\mathcal{P}}$ if $f(\mu)$ is either 0 or $*$ for all $\mu \in \mathcal{P}$; similarly, $f$ *is equivalent to* $\top_{D^*_\mathcal{P}}$ if $f(\mu)$ is either 1 or $*$ for all $\mu \in \mathcal{P}$. (Since, by definition, $f(\mu) \neq *$ for at least one $\mu \in \mathcal{P}$, an element cannot be equivalent to both $\top_{D^*_\mathcal{P}}$ and $\bot_{D^*_\mathcal{P}}$.) Let $D^*_\mathcal{P}$ be the same as $D'_\mathcal{P}$ except that all elements equivalent to $\bot_{D_\mathcal{P}}$ are identified (and viewed as one element) and all elements equivalent to $\top_{D^*_\mathcal{P}}$ are identified. More precisely, let $D^*_\mathcal{P} = \{\bot_{D^*_\mathcal{P}}, \top_{D^*_\mathcal{P}}\} \cup \{f \in D' : f \text{ is not equivalent to } \top_{D^*_\mathcal{P}} \text{ or } \bot_{D^*_\mathcal{P}}\}$. Define the ordering $\leq$ on $D^*_\mathcal{P}$ by taking $f \leq g$ if one of the following three conditions holds:

- $f = \bot_{D^*_\mathcal{P}}$,
- $g = \top_{D^*_\mathcal{P}}$,
- neither $f$ nor $g$ is $\bot_{D^*_\mathcal{P}}$ or $\top_{D^*_\mathcal{P}}$ and for all $\mu \in \mathcal{P}$, either $f(\mu) = g(\mu) = *$ or $f(\mu) \neq *$, $g(\mu) \neq *$, and $f(\mu) \leq g(\mu)$.

Now define

$$\text{Pl}_\mathcal{P}(U|V) = \begin{cases} \bot_{D^*_\mathcal{P}} & \text{if } \exists \mu \in \mathcal{P}(\mu(V) \neq 0) \text{ and} \\ & \forall \mu \in \mathcal{P}(\mu(V) \neq 0 \Rightarrow \mu(U|V) = 0), \\ \top_{D^*_\mathcal{P}} & \text{if } \exists \mu \in \mathcal{P}(\mu(V) \neq 0) \text{ and} \\ & \forall \mu \in \mathcal{P}(\mu(V) \neq 0 \Rightarrow \mu(U|V) = 1), \\ \text{und.} & \text{if } \mu(V) = 0 \text{ for all } \mu \in \mathcal{P}, \\ f_{U|V} & \text{otherwise.} \end{cases}$$

It is easy to check that this gives a coherent cps. I remark that a similar construction can be used to convert any FH-cps to a cps and contruct a conditional plausibility measure from an unconditional plausibility measure. I leave details to the full paper.

These constructions for extending an unconditional measure of likelihood to a cps have two properties that are worth abstracting. A cps $(W, \mathcal{F}, \mathcal{F}', \text{Pl})$ is *standard* if $\mathcal{F}' = \{U : \text{Pl}(U) \neq \bot\}$. Note that all the constructions of cps's given above result in standard cps's. This follows from a more general observation. $(W, \mathcal{F}, \mathcal{F}', \text{Pl})$ is *determined by unconditional plausibility* if there is a function $g$ such that $\text{Pl}(U|V) = g(\text{Pl}(U \cap V|W), \text{Pl}(V|W))$ for all $(U, V) \in \mathcal{F} \times \mathcal{F}'$. It is almost immediate from the definitions that all the constructions above result in cps's that are determined by unconditional plausibility. If an acceptable cps is determined by unconditional plausibility, then it must be standard.

**Lemma 2.1:** *If $(W, \mathcal{F}, \mathcal{F}', \text{Pl})$ is an acceptable cps determined by unconditional plausibility such that $\text{Pl}(W) \neq \text{Pl}(\emptyset)$, then $(W, \mathcal{F}, \mathcal{F}', \text{Pl})$ is a standard cps.*

## 3 ALGEBRAIC CONDITIONAL PLAUSIBILITY MEASURES

To be able to carry out the type of reasoning used in Bayesian networks, it does not suffice to just have conditional plausibility. We need to have analogues of addition and multiplication. More precisely, there needs to be some way of computing the plausibility of the union of two disjoint sets in terms of the plausibility of the individual sets and a way of computing $\text{Pl}(U \cap V|V')$ given $\text{Pl}(U|V \cap V')$ and $\text{Pl}(V|V')$.

A cps $(W, \mathcal{F}, \mathcal{F}', \text{Pl})$ where Pl has range $D$ is *algebraic* if it is acceptable and there are functions $\oplus : D \times D \to D$ and $\otimes : D \times D \to D$ such that the following properties hold:

Alg1. If $U, U' \in \mathcal{F}$ are disjoint and $V \in \mathcal{F}'$, then $\text{Pl}(U \cup U'|V) = \text{Pl}(U|V) \oplus \text{Pl}(U'|V)$.

Alg2. If $U \in \mathcal{F}$, $V \cap V' \in \mathcal{F}'$, then $\text{Pl}(U \cap V|V') = \text{Pl}(U|V \cap V') \otimes \text{Pl}(V|V')$.

Alg3. $\otimes$ distributes over $\oplus$, more precisely, $a \otimes (b \oplus b') = (a \otimes b) \oplus (a \otimes b')$ if $(a, b), (a, b'), (a, b \oplus b') \in Dom(\otimes)$ and $(b, b'), (a \otimes b, a \otimes b') \in Dom(\oplus))$, where $Dom(\oplus) = \{(\text{Pl}(U|V), \text{Pl}(U'|V)) : U, U' \in \mathcal{F}$ are disjoint and $V \in \mathcal{F}'\}$ and $Dom(\otimes) = \{(\text{Pl}(U|V \cap V'), \text{Pl}(V|V')) : U \in \mathcal{F}, V \cap V' \in F'\}$.

Alg4. If $(a, c), (b, c) \in Dom(\otimes)$, $a \otimes c \leq b \otimes c$, and $c \neq \bot$, then $a \leq b$.

I sometimes refer to the cpm Pl as being algebraic as well.

It may seem more natural to consider a stronger version of Alg4 that applies to all pairs in $D \times D$, such as

Alg4′. If $a \otimes c \leq b \otimes c$ and $c \neq \bot$, then $a \leq b$.

However, as Proposition 3.1 below shows, by requiring that Alg3 and Alg4 hold only for tuples in $Dom(\oplus)$ and $Dom(\otimes)$ rather than on all tuples in $D \times D$, some cps's of interest become algebraic that would otherwise not be. Intuitively, we care about $\oplus$ and $\otimes$ mainly to the extent that Alg1 and Alg2 hold, and they apply to tuples in $Dom(\oplus)$ and $Dom(\otimes)$, respectively. Thus, it does not seem unreasonable that properties like Alg3 and Alg4 be required to hold only for these tuples.

**Proposition 3.1:** *The constructions for extending an unconditional probability measure, ranking function, possibility measure (using either $\text{Poss}(U|V)$ or $\text{Poss}(U\|V)$), and the plausibility measure $\text{Pl}_\mathcal{P}$ defined by a set $\mathcal{P}$ of probability measures to a cps result in algebraic cps's.*[1]

**Proof:** It is easy to see that in each case the cps is acceptable. It is also easy to find appropriate notions of $\otimes$ and $\oplus$ in the case of probability measures, ranking functions, and possibility measures using $\text{Poss}(U|V)$. For probability, clearly $\oplus$ and $\otimes$ are $+$ and $\times$ (more precisely, $a \oplus b = \min(1, a + b)$); for ranking, $\oplus$ and $\otimes$ are min and $+$; for $\text{Poss}(U\|V)$, $\oplus$ is max and $\otimes$ is $\times$. I leave it to the reader to check that Alg1–4 hold in all these cases.

For $\text{Poss}(U|V)$, $\oplus$ is again max and $\otimes$ is min. Note that if $(a, b) \in Dom(\min)$, then either $a < b$ or $a = 1$. For suppose that $(a, b) = (\text{Poss}(U|V \cap V'), \text{Poss}(V|V'))$, where $U \in \mathcal{F}$ and $V \cap V' \in \mathcal{F}'$. If $\text{Poss}(U \cap V \cap V') = \text{Poss}(V \cap V')$ then $a = \text{Poss}(U|V \cap V') = 1$; otherwise, $\text{Poss}(U \cap V \cap V') < \text{Poss}(V \cap V')$, in which case $a = \text{Poss}(U|V \cap V') = \text{Poss}(U \cap V \cap V') < \text{Poss}(V \cap V') \leq \text{Poss}(V|V') = b$. It is easy to check Alg1–3. While min does not satisfy Alg4′—certainly $\min(a, c) = \min(b, c)$ does not in general imply that $a = b$—Alg4 does hold. For if $\min(a, c) \leq \min(b, c)$ and $a = 1$, then clearly $b = 1$. Alternatively, if $a < c$, then $\min(a, c) = a$ and the only way that $a \leq \min(b, c)$, given that $b < c$ or $b = 1$, is if $a \leq b$.

Finally, for $\text{Pl}_\mathcal{P}$, $\oplus$ and $\otimes$ are essentially pointwise addition and multiplication. However, we must be a little careful in dealing with $\top_{D_\mathcal{P}^*}$, $\bot_{D_\mathcal{P}^*}$, and $*$. The definition of $\oplus$ is relatively straightforward. Define $f \oplus \top_{D_\mathcal{P}^*} = \top_{D_\mathcal{P}^*} \oplus f = \top_{D_\mathcal{P}^*}$ and $f \oplus \bot_{D_\mathcal{P}^*} = \bot_{D_\mathcal{P}^*} \oplus f = f$. If $f, g \cap \{\bot_{D_\mathcal{P}^*}, \top_{D_\mathcal{P}^*}\} = \emptyset$, then $f \oplus g = h$, where $h(\mu) = \min(1, f(\mu) + g(\mu))$ (taking $a + * = * + a = *$ and $\min(1, *) = *$). In a similar spirit, define $f \otimes \top_{D_\mathcal{P}^*} = \top_{D_\mathcal{P}^*} \otimes f = f$ and $f \otimes \bot_{D_\mathcal{P}^*} = \bot_{D_\mathcal{P}^*} \otimes f = \bot_{D_\mathcal{P}^*}$; if $\{f, g\} \cap \{\bot_{D_\mathcal{P}^*}, \top_{D_\mathcal{P}^*}\} = \emptyset$, then $f \otimes g = h$, where $h(\mu) = f(\mu) \times g(\mu)$ (taking $* \times a = a \times * = *$). I leave it to the reader to check that, with these definitions, Alg1–4 hold (although note that the restrictions to $Dom(\oplus)$ and $Dom(\otimes)$ are required for both Alg3 and Alg4 to hold). ∎

Conditional belief and (conditional) lower probability are not algebraic. There is no analogue to either $\oplus$ or $\otimes$. For example, in the case of lower probability, it is not hard to construct pairwise disjoint sets $U_1$, $V_1$, $U_2$, and $V_2$ and a set $\mathcal{P}$ of probability measures such that $\mathcal{P}_*(U_i) = \mathcal{P}_*(V_i)$ (and $\mathcal{P}^*(U_i) = \mathcal{P}^*(V_i)$) for $i = 1, 2$, but $\mathcal{P}_*(U_1 \cup U_2) \neq \mathcal{P}_*(V_1 \cup V_2)$. That means there cannot be a function $\oplus$ in the case of lower probability. Similar remarks hold for $\otimes$ and for belief functions.

I conclude this section by showing that a standard algebraic cps that satisfies one other minimal property must also satisfy CPl5. Say that $\otimes$ is *monotonic* if $d \leq d'$ and $e \leq e'$ then $d \otimes e \leq d' \otimes e'$. A cpm (cps) is monotonic if $\otimes$ is.

---

[1] Essentially the same result is proved in [Friedman and Halpern 1995] for all cases but $\text{Pl}_\mathcal{P}$.

**Lemma 3.2:** *A standard algebraic monotonic cps satisfies CPI5.*

## 4 INDEPENDENCE

How can we capture formally the notion that two events are *independent*? Intuitively, it means that they have nothing to do with each other—they are totally unrelated; the occurrence of one has no influence on the other. None of the representations of uncertainty that we have been considering can express the notion of "unrelatedness" (whatever it might mean) directly. The best we can do is to capture the "footprint" of independence on the notion. For example, in the case of probability, if $U$ and $V$ are unrelated, it seems reasonable to expect that learning $U$ should not affect the probability of $V$ and symmetrically, learning $V$ should not affect the probability of $U$. "Unrelatedness" is, after all, a symmetric notion.[2] The fact that $U$ and $V$ are probabilistically independent (with respect to probability measure $\mu$) can thus be expressed as $\mu(U|V) = \mu(U)$ and $\mu(V|U) = \mu(V)$. There is a technical problem with this definition: What happens if $\mu(V) = 0$? In that case $\mu(U|V)$ is undefined. Similarly, if $\mu(U) = 0$ then $\mu(V|U)$ is undefined. It is conventional to say that, in this case, $U$ and $V$ are still independent. This leads to the following formal definition.

**Definition 4.1:** $U$ and $V$ are *probabilistically independent (with respect to probability measure $\mu$)* if $\mu(V) \neq 0$ implies $\mu(U|V) = \mu(U)$ and $\mu(U) \neq 0$ implies $\mu(V|U) = \mu(V)$. ∎

This does not look like the standard definition of independence in texts, but an easy calculation shows that it is equivalent.

**Proposition 4.2:** *The following are equivalent:*

(a) $\mu(U) \neq 0$ *implies* $\mu(V|U) = \mu(V)$,

(b) $\mu(U \cap V) = \mu(U)\mu(V)$,

(c) $\mu(V) \neq 0$ *implies* $\mu(U|V) = \mu(U)$.

Thus, in the case of probability, it would be equivalent to say that $U$ and $V$ are independent with respect to $\mu$ if $\mu(U \cap V) = \mu(U)\mu(V)$ or to require only that $\mu(U|V) = \mu(U)$ if $\mu(V) \neq 0$ without requiring that $\mu(V|U) = \mu(V)$ if $\mu(U) \neq 0$. However, these equivalences do not necessarily hold for other representations of uncertainty. The definition of independence I have given here seems to generalize more appropriately.[3]

The definition of probabilistic conditional independence is analogous.

**Definition 4.3:** $U$ and $V$ are *probabilistically independent given $V'$ (with respect to probability measure $\mu$)* if $\mu(V \cap V') \neq 0$ implies $\mu(U|V \cap V') = \mu(U|V')$ and $\mu(U \cap V') \neq 0$ implies $\mu(V|U \cap V') = \mu(V|V')$. ∎

It is immediate that $U$ and $V$ are (probabilistically) independent iff they are independent conditional on $W$.

The generalization to conditional plausibility measures (and hence to all other representations of uncertainty that we have been considering) is straightforward.

**Definition 4.4:** Given a cps $(W, \mathcal{F}, \mathcal{F}', \text{Pl})$, $U, V \in \mathcal{F}$ are *plausibilistically independent given $V' \in \mathcal{F}$ (with respect to the cpm Pl)*, written $I_{\text{Pl}}(U, V|V')$, if $V \cap V' \in \mathcal{F}'$ implies $\text{Pl}(U|V \cap V') = \text{Pl}(U|V')$ and $U \cap V' \in \mathcal{F}'$ implies $\text{Pl}(V|U \cap V') = \text{Pl}(V|V')$. ∎

We are interested in conditional independence of random variables as well as in conditional independence of events. All the standard definitions extend to plausibility in a straightforward way. As usual, a *random variable* $X$ on $W$ is a function from $W$ to the reals. Let $\mathcal{R}(X)$ be the set of possible values for $X$ (that is, the set of values over which $X$ ranges). As usual, $X = x$ is the event $\{w : X(w) = x\}$. If $\mathbf{X} = \{X_1, \ldots, X_k\}$ is a set of random variables and $\mathbf{x} = (x_1, \ldots, x_k)$, let $\mathbf{X} = \mathbf{x}$ be an abbreviation for the event $X_1 = x_1 \cap \ldots \cap X_k \cap x_k$. A random variable is *measurable* with respect to cps $(W, \mathcal{F}, \mathcal{F}', \text{Pl})$ if $X = x \in \mathcal{F}$ for all $x \in \mathcal{R}(X)$. For the rest of the paper, I assume that all random variables $X$ are measurable and that $\mathcal{R}(X)$ is finite for all random variables $X$. Random variables $X$ and $Y$ are independent with respect to plausibility measure Pl if the events $X = x$ and $Y = y$ are independent for all $x \in \mathcal{R}(X)$ and $y \in \mathcal{R}(Y)$. More generally, given sets $\mathbf{X}$, $\mathbf{Y}$, and $\mathbf{Z}$ of random variables, $\mathbf{X}$ and $\mathbf{Y}$ are plausibilistically independent given $\mathbf{Z}$ (with respect to Pl), denoted $I_{\text{Pl}}^{rv}(\mathbf{X}, \mathbf{Y}|\mathbf{Z})$, if $I_{\text{Pl}}(\mathbf{X} = \mathbf{x}, \mathbf{Y} = \mathbf{x}|\mathbf{Z} = \mathbf{z})$ for all $\mathbf{x}$, $\mathbf{y}$, and $\mathbf{z}$. (Note that I am using $I_{\text{Pl}}$ for conditional independence of events and $I_{\text{Pl}}^{rv}$ for conditional independence of random variables.) If $\mathbf{Z} = \emptyset$, then $I_{\text{Pl}}^{rv}(\mathbf{X}, \mathbf{Y}|\mathbf{Z})$ if $\mathbf{X}$ and $\mathbf{Y}$ are unconditionally independent, that is, if $I_{\text{Pl}}(\mathbf{X} = \mathbf{x}, \mathbf{Y} = \mathbf{x}|W)$ for all $\mathbf{x}$, $\mathbf{y}$; if either $\mathbf{X} = \emptyset$ or $\mathbf{Y} = \emptyset$, then $I^{rv}(\mathbf{X}, \mathbf{Y}|\mathbf{Z})$ is taken to be vacuously true.

Now consider the following four properties of random variables, called the *semi-graphoid properties* [Pearl 1988], where $\mathbf{X}$, $\mathbf{Y}$, and $\mathbf{Z}$ are pairwise disjoint sets of variables.

CIRV1. If $I_{\text{Pl}}^{rv}(\mathbf{X}, \mathbf{Y}|\mathbf{Z})$ then $I_{\text{Pl}}^{rv}(\mathbf{Y}, \mathbf{X}|\mathbf{Z})$.

CIRV2. If $I_{\text{Pl}}^{rv}(\mathbf{X}, \mathbf{Y} \cup \mathbf{Y}'|\mathbf{Z})$ then $I_{\text{Pl}}^{rv}(\mathbf{X}, \mathbf{Y}|\mathbf{Z})$.

CIRV3. If $I_{\text{Pl}}^{rv}(\mathbf{X}, \mathbf{Y} \cup \mathbf{Y}'|\mathbf{Z})$ then $I_{\text{Pl}}^{rv}(\mathbf{X}, \mathbf{Y}|\mathbf{Y}' \cup \mathbf{Z})$.

CIRV4. If $I_{\text{Pl}}^{rv}(\mathbf{X}, \mathbf{Y}|\mathbf{Z})$ and $I_{\text{Pl}}^{rv}(\mathbf{X}, \mathbf{Y}'|\mathbf{Y} \cup \mathbf{Z})$ then $I_{\text{Pl}}^{rv}(\mathbf{X}, \mathbf{Y} \cup \mathbf{Y}'|\mathbf{Z})$.

---

[2] Walley [1991] calls the asymmetric notion *irrelevance* and defines $U$ being independent of $V$ as $U$ is irrelevant to $V$ and $V$ is irrelevant to $U$. Although my focus here is independence, I do not mean to suggest that irrelevance is uninteresting.

[3] Another property of probabilistic independence is that if $U$ is independent of $V$ then $\overline{U}$ is independent of $V$. This too does not follow for the other representations of uncertainty, and Walley [1991] actually makes this part of his definition. Adding this

requirement would not affect any of the results here, although it would make the proofs somewhat lengthier, so I have not made it part of the definition.

It is well known that CIRV1–4 hold for probability measures. The following result generalizes this. The proof is not difficult, although care must be taken to show that the result depends only on the properties of algebraic cpm's.

**Theorem 4.5:** *CIRV1–4 hold for all algebraic cps's.*

Theorem 4.5, of course, is very dependent on the definition of conditional independence given here. Other notions of independence have been studied in the literature for specific representations of uncertainty. There is a general approach called *noninteractivity*, which was originally defined in the context of possibility measures by Zadeh [1978] but makes sense for any algebraic cpm. $U$ and $V$ *do not interact given* $V'$ *(with respect to Pl)*, denoted $NI_{\text{Pl}}(U,V|V')$ if $V' \in \mathcal{F}'$ implies that $\text{Pl}(U \cap V|V') = \text{Pl}(U|V') \otimes \text{Pl}(V|V')$.[4] Fonck [1994] shows that noninteraction is strictly weaker than independence for a number of notions of independence for possibility measures. The following result shows that noninteraction implies independence for all algebraic cpm's.

**Lemma 4.6:** *If* $(W, \mathcal{F}, \mathcal{F}', Pl)$ *is an algebraic cps, then* $I_{\text{Pl}}(U,V|V')$ *implies* $NI_{\text{Pl}}(U,V|V')$.

What about the converse to Lemma 4.6? The results of Fonck show that it does not hold in general—indeed, it does not hold for Poss($U|V$). So what is required for noninteractivity to imply independence? The following lemma provides a sufficient condition.

**Lemma 4.7:** *If* $(W, \mathcal{F}, \mathcal{F}', Pl)$ *is a standard algebraic cps that satisfies Alg4', then* $NI_{\text{Pl}}(U,V|V')$ *implies* $I_{\text{Pl}}(U,V|V')$.

The fact that noninteractivity and conditional independence coincide for the conditional plausibility spaces constructed from unconditional probability measures, ranking functions, and possibility measures using Poss($U||V$) follows from Lemmas 4.6 and 4.7. Since neither Poss($U|V$) nor Pl$_\mathcal{P}$ satisfy Alg4', it is perhaps not surprising that in neither case does noninteractivity imply conditional independence. (We shall shortly see an example in the case of Pl$_\mathcal{P}$; Fonck [1994] gives examples in the case of Poss($U|V$).)

It is easy to see that the assumption of standardness is necessary in Lemma 4.7. For suppose that $(W, \mathcal{F}, \mathcal{F}', \text{Pl})$ is a nonstandard algebraic cps for which $\top \neq \bot$. Since $(W, \mathcal{F}, \mathcal{F}', \text{Pl})$ is nonstandard, there must exist some $U \in \mathcal{F}'$ such that $\text{Pl}(U|W) = \bot$. But then

$$\bot = \text{Pl}(\emptyset|W) = \text{Pl}(\emptyset|U) \otimes \text{Pl}(U|W) = \bot \otimes \bot.$$

Thus

$$\text{Pl}(U|W) = \bot = \bot \otimes \bot = \text{Pl}(U|W) \otimes \text{Pl}(U|W),$$

so $NI_{\text{Pl}}(U,U|W)$. But $\text{Pl}(U|U) = \top \neq \bot = \text{Pl}(U)$, so $I_{\text{Pl}}(U,U|W)$ does not hold.

In general, Theorem 4.5 does not hold if we use $NI_{\text{Pl}}$ rather than $I_{\text{Pl}}$. Besides noninteractivity, a number of different approaches to defining independence for possibility measures [Campos and Huete 1999a; Campos and Huete 1999b; Dubois, Farinas del Cerro, Herzig, and Prade 1994] and for sets of probability measures [Campos and Huete 1993; Campos and Moral 1995; Couso, Moral, and Walley 1999] have been considered. In general, Theorem 4.5 does not hold for them either. It is beyond the scope of this paper to discuss and compare these approaches to that considered here, but it is instructive to consider independence for sets of probability measures in a little more detail, especially for the representation Pl$_\mathcal{P}$.

$I_{\text{Pl}_\mathcal{P}}$ is very close to a notion called *type-1* independence considered by de Campos and Moral [1995]. $U$ and $V$ are *type-1 independent conditional on* $V'$ *with respect to* $\mathcal{P}$ if $U$ and $V$ are independent conditional on $V'$ with respect to every $\mu \in \mathcal{P}$. It is easy to check that $I_{\text{Pl}_\mathcal{P}}(U,V|V')$ implies that $U$ and $V$ are type-1 independent conditional on $V'$ (and similarly for random variables); however, the converse does not necessarily hold, because the two approaches treat conditioning on events that have probability 0 according to some (but not all) of the measures in $\mathcal{P}$ differently. To see this, consider an example discussed by de Campos and Moral. Suppose a coin is known to be either double-headed or double-tailed and is tossed twice. This can be represented by $\mathcal{P} = \{\mu_0, \mu_1\}$, where $\mu_0(hh) = 1$ and $\mu_0(ht) = \mu_0(th) = \mu_0(tt) = 0$, while $\mu_1(tt) = 1$ and $\mu_1(ht) = \mu_1(th) = \mu(hh) = 0$. Let $X_1$ and $X_2$ be the random variables representing the outcome of the first and second coin tosses, respectively. Clearly there is a functional dependence between $X_1$ and $X_2$, but it is easy to check that $X_1$ and $X_2$ are type-1 independent with respect to $\mathcal{P}$. Moreover, noninteractivity holds: $NI_{\text{Pl}}(X_1 = i, X_2 = j)$ holds for $i,j \in \{h,t\}$. On the other hand, $I_{\text{Pl}_\mathcal{P}}(X_1, X_2)$ does not hold. For example, $f_{X_1=h}(\mu_1) = 0$ while $f_{X_1=h|X_2=h}(\mu_1) = *$.

## 5  BAYESIAN NETWORKS

Throughout this section, I assume that we start with a set $W$ of possible worlds characterized by $n$ binary random variables $\mathcal{X} = \{X_1, \ldots, X_n\}$ (or, equivalently, $n$ primitive propositions). That is, a world $w \in W$ is a tuple $(x_1, \ldots, x_n)$, where $x_i \in \{0,1\}$ is the value of $X_i$. That means that there are $2^n$ worlds in $W$, say $w_1, \ldots, w_{2^n}$.[5] Let $\mathcal{PL}$ consist of all algebraic cps's of the form $(W, \mathcal{F}, \mathcal{F}', \text{Pl})$ determined by unconditional plausibility by some fixed function $g$, where $\mathcal{F} = 2^W$, so that all subsets of $W$ are measurable. Thus, for example, $\mathcal{PL}$ could consist of any of the families of conditional plausibility measures constructed in Section 2. If $(W, \mathcal{F}, \mathcal{F}', \text{Pl}) \in \mathcal{PL}$, the cpm Pl can be described by giving Pl($w$) for each of the $2^n$ worlds $w$ in $W$. (Since Pl is algebraic, the unconditional plausibility of a set of worlds is determined by the unconditional plausibility of individual worlds in the set; since Pl is determined by unconditional plausibility, this suffices to determine all conditional plausibilities.) The goal of this section is to show that many of the tools of Bayesian network technology can be applied in this setting. Almost all the results follow easily from Theorem 4.5 and well-known results in the literature (mainly in [Geiger and Pearl 1988; Geiger, Verma, and Pearl 1990; Verma 1986]), so I just sketch the details here.

---

[4]Shenoy [1994] defines a notion similar in spirit to noninteractivity for random variables.

[5]The assumption that the random variables are binary is just for ease of exposition. It is easy to generalize the results to the case where $\mathcal{R}(X_i)$ is finite for each $X_i$.

Because Pl is algebraic, the chain rule holds. In particular, it follows from Alg2 that

$$\text{Pl}(X_1 = x_1 \cap \ldots \cap X_n = x_n) =$$
$$\text{Pl}(X_n = x_n | X_1 = x_1 \cap \ldots \cap X_{n-1} = x_{n-1}) \otimes$$
$$\text{Pl}(X_{n-1} = x_{n-1} | X_1 = x_1 \cap \ldots \cap X_{n-2} = x_{n-2}) \otimes$$
$$\ldots \otimes \text{Pl}(X_2 = x_2 | X_1 = x_1) \otimes \text{Pl}(X_1 = x_1). \quad (1)$$

Strictly speaking, I should put in parentheses here, since nothing in Alg1–4 forces $\otimes$ to be associative. However, it follows easily from Alg2 that $\otimes$ is in fact associative on tuples $(a, b, c)$ of the form $(\text{Pl}(U_1|U_2), \text{Pl}(U_2|U_3), \text{Pl}(U_3|U_4))$, where $U_1 \subseteq U_2 \subseteq U_3 \subseteq U_4$ (see also [Friedman and Halpern 1995]), which are the only types of tuples that arise in (1), so the parentheses can safely be omitted.

As usual, a (qualitative) *Bayesian network* (over $\mathcal{X}$) is a *dag* whose nodes are labeled by variables in $\mathcal{X}$. The standard notion of a Bayesian network representing a probability measure [Pearl 1988] can be generalized in the obvious way to plausibility.

**Definition 5.1:** Given a qualitative Bayesian network $G$, let $\text{Par}_G(X)$ be the *parents* of the random variable $X$ in $G$; let $\text{Des}_G(X)$ be all the *descendants* of $X$, that is, $X$ and all those nodes $Y$ such that $X$ is an ancestor of $Y$; let $\text{NonDes}_G(X)$, the *nondescendants of* $X$, consist of $\mathcal{X} - \text{Des}_G(X)$. Note that all ancestors of $X$ are nondescendants of $X$. The Bayesian network $G$ *is compatible with* Pl if $I_{\text{Pl}}^{rv}(X, \text{NonDes}_G(X) | \text{Par}(X))$, that is, $X$ is conditionally independent of its nondescendants given its parents, for all $X \in \mathcal{X}$. ∎

A *quantitative Bayesian network* is a pair $(G, f)$ consisting of a qualitative Bayesian network $G$ and a function $f$ that associates with each node $X$ in $G$ a *conditional plausibility table (cpt)* that quantifies the effects of the parents of $X$ on $X$. There is an entry in $D$ (the range of Pl) in the cpt for each possible setting of the parents of $X$. Intuitively, the entries in the cpt for $X$ describe the plausibility that $X = 1$ conditional on all the possible values of $X$'s parents. If $X$ is a root of $G$, then the cpt for $X$ can be thought of as giving the unconditional plausibility that $X = 1$.

**Definition 5.2:** A quantitative Bayesian network $(G, f)$ *is compatible with* Pl if $G$ qualitatively represents $\mu$ and the cpts agree with Pl, in the sense that, for each random variable $X$, the entry in the cpt for $X$ given some setting $Y_1 = y_1, \ldots, Y_k = y_k$ of its parents is $\text{Pl}(X = 1 | Y_1 = y_1 \cap \ldots \cap Y_k = y_k)$ if $Y_1 = y_1 \cap \ldots \cap Y_k = y_k \in \mathcal{F}'$. (It does not matter what the cpt entry for $Y_1 = y_1, \ldots, Y_k = y_k$ is if $Y_1 = y_1 \cap \ldots \cap Y_k = y_k \notin \mathcal{F}'$.) ∎

It follows from (1) and Alg1 that Pl can be reconstructed from a quantitative Bayesian network representing Pl.

There is a standard way of constructing a Bayesian network that represents a probability measure [Pearl 1988]. Let $Y_1, \ldots, Y_n$ be a permutation of the random variables in $\mathcal{X}$. Construct a qualitative Bayesian network as follows: For each $k$, find a minimal subset of $\{Y_1, \ldots, Y_{k-1}\}$, call it $\mathbf{P}_k$, such that $I_{\text{Pl}}^{rv}(\{Y_1, \ldots, Y_{k-1}\}, Y_k | \mathbf{P}_k)$. Then add edges from each of the nodes in $\mathbf{P}_k$ to $Y_k$. Call the resulting graph $G$. Verma [1986] shows that this construction gives a Bayesian network compatible with Pl in the case of probability; his proof depends only on CIRV1–4. Thus, the construction works for algebraic cpm's.

**Theorem 5.3:** *If Pl is an algebraic cpm, then $G$ is compatible with Pl.*

Just as in the case of probability, conditional independencies can be read off the Bayesian network using the criterion of d-separation [Pearl 1988]. Recall that a set $\mathbf{X}$ of nodes in $G$ is *d-separated* from a set $\mathbf{Y}$ of nodes by a set $\mathbf{Z}$ of nodes if, for every $X \in \mathbf{X}, Y \in \mathbf{Y}$, and *undirected path* from $X$ to $Y$ (an undirected path is a path that ignores the arrows), there is a node $Z'$ on the path such that either:

(a) $Z' \in \mathbf{Z}$ and there is an arrow on the path leading in to $Z'$ and an arrow leading out;

(b) $Z' \in \mathbf{Z}$ and has both path arrows leading out; or

(c) $Z'$ has both path arrows leading in, and neither $Z'$ nor any of its descendants are in $\mathbf{Z}$.

Let $\Sigma_{G,\text{Pl}}$ consist of all statements of the form $I_{\text{Pl}}^{rv}(\text{NonDes}_G(X), X | \text{Par}_G(X))$. Consider the following three statements:

1. $d\text{-}sep_G(\mathbf{X}, \mathbf{Y} | \mathbf{Z})$.

2. $I_{\text{Pl}}^{rv}(\mathbf{X}, \mathbf{Y} | \mathbf{Z})$ is provable from CIRV1–4 and $\Sigma_{G,\text{Pl}}$.

3. $I_{\text{Pl}}^{rv}(\mathbf{X}, \mathbf{Y} | \mathbf{Z})$ holds for every plausibility measure in $\mathcal{PL}$ compatible with $G$.

The implication from 1 to 2 is proved in [Geiger, Verma, and Pearl 1990; Verma 1986].

**Theorem 5.4:** [Geiger, Verma, and Pearl 1990; Verma 1986] *If $d\text{-}sep_G(\mathbf{X}, \mathbf{Y} | \mathbf{Z})$, then $I_{\text{Pl}}^{rv}(\mathbf{X}, \mathbf{Y} | \mathbf{Z})$ is provable from CIRV1–4 and $\Sigma_{G,\text{Pl}}$.*

It is immediate from Theorem 4.5 that the implication from 2 to 3 holds for algebraic cpm's.

**Corollary 5.5:** *If $I_{\text{Pl}}^{rv}(\mathbf{X}, \mathbf{Y} | \mathbf{Z})$ is provable from CIRV1–4 and $\Sigma_{G,\text{Pl}}$, then $I_{\text{Pl}}^{rv}(\mathbf{X}, \mathbf{Y} | \mathbf{Z})$ holds for every algebraic cpm Pl compatible with $G$.*

Finally, the implication from 3 to 1 for probability measures is proved in [Geiger and Pearl 1988; Geiger, Verma, and Pearl 1990]. Here I generalize the proof to algebraic plausibility measures. Notice that to prove the implication from 3 to 1, it suffices to show that if $X$ is not d-separated from $Y$ by $\mathbf{Z}$ in $G$, then there is a plausibility measure $\text{Pl} \in \mathcal{PL}$ such that $I_{\text{Pl}}^{rv}(X, Y | \mathbf{Z})$ does not hold. To guarantee that such a plausibility measure exists in $\mathcal{PL}$, we have to ensure that there are "enough" plausibility measures in $\mathcal{PL}$ in the following technical sense. $\mathcal{PL}$ is *rich* if for all pairs $(d, d') \in D$ such that $d \leq d' \neq \bot_D$, there exists $\text{Pl}_{d,d'} \in \mathcal{PL}$ and sets $U, V \subseteq W$ such that $\text{Pl}(U \cap V | W) = d$ and $\text{Pl}(V | W) = d'$. All the constructions given in Section 2 for constructing cps's from unconditional measures of likelihood result in rich cps's.

**Theorem 5.6** *Suppose that $\mathcal{PL}$ is rich. Then if $I_{\text{Pl}}^{\text{IV}}(\mathbf{X}, \mathbf{Y} | \mathbf{Z})$ holds for every plausibility measure in $\mathcal{PL}$ compatible with $G$, then $d\text{-}sep_G(\mathbf{X}, \mathbf{Y} | \mathbf{Z})$.*

## 6 CONCLUSION

I have considered a general notion of conditional plausibility that generalizes all other standard notions of conditioning in the literature, and examined various requirements that could be imposed on conditional plausibility. One set of requirements, those that lead to algebraic cps's, was shown to suffice for the construction of Bayesian networks. It should also be clear that standard constructions like belief propagation in Bayesian networks [Pearl 1988] can also be applied to algebraic cps's, since they typically use only basic properties of conditioning, addition, and multiplication, all of which hold in algebraic cps's (using $\oplus$ and $\otimes$). In particular, these results apply to sets to probability measures, provided that they are appropriately represented as plausibility measures. The particular representation of sets of probability measures advocated in this paper was also shown to have a number of other attractive properties.

### Acknowledgments

I thank Serafín Moral, Fabio Cozman, and the anonymous referees for very useful comments. This work was supported in part by the NSF, under grant IRI-96-25901.